# Automated Monitoring Cropland Using Remote Sensing Data: Challenges and Opportunities for Machine Learning

Xiaowei Jia, Ankush Khandelwal, Vipin Kumar

*Department of Computer Science and Engineering, University of Minnesota, Twin Cities, US*

**Abstract**

This paper provides an overview of how recent advances in machine learning and the availability of data from earth observing satellites can dramatically improve our ability to automatically map croplands over long period and over large regions. It discusses three applications in the domain of crop monitoring where ML approaches are beginning to show great promise. For each application, it highlights machine learning challenges, proposed approaches, and recent results. The paper concludes with discussion of major challenges that need to be addressed before ML approaches will reach their full potential for this problem of great societal relevance.

# 1. Introduction

Agriculture is one of the most crucial ecosystems for human sustenance. Agricultural resources are witnessing tremendous supply side stresses as a result of rapidly increasing population, sub-optimal farming practices, increased pest damage occurrences due to climate change, and loss of productive land by other human activities such as urbanization (Ortiz et al., 2008; d'Amour et al., 2017; Bebber et al., 2014). Hence, multi-agency international initiatives such as GEO-GLAM (Singh et al., 2012) and Global Yield Gap Atlas (van Ittersum et al., 2018) have been created with the goal of improving our ability to produce and share relevant, timely and accurate trends and forecasts of crop productivity globally. Efforts within these initiatives have historically depended on surveys and self-reported statistics at regional and national scale and thus are necessarily limited in scope and resolution both in space and time.

Advances in Earth observation technologies have led to the acquisition of vast amounts of Earth system data that can be used for monitoring changes on a global scale. In particular, a wide variety of instruments and sensors onboard satellites operated by the US and other international agencies collect petabytes of data on a regular basis. For example, MODIS sensors onboard Terra and Aqua satellites have been collecting optical data daily at 500m spatial resolution since 2000. Another commonly used dataset is from the Landsat series of satellites at bi-weekly temporal scale and 30m spatial resolution since 1974. More recently, the Sentinel series of satellites launched by the European Space Agency have been capturing data both through optical and radar sensors with a spatial resolution of at least 10m and temporal resolution of 5 to 10 days. The advent of microsatellite constellations for use in earth observation, such as those offered by Planet, is going to further increase the depth and breadth of remotely sensed data assets. These rich datasets hold great potential for cropland monitoring as they contain rich temporal and spectral information to map different types of crops and estimate the changes in spatial distribution of different crops around the world due to climate, market and policy changes (Alston et al., 2010, Beddow et al., 2015).

Machine learning models, which have found tremendous success in commercial applications e.g., computer vision and natural language processing (Liu et al., 2017), are increasing being considered as alternatives to surveys and self-reported data to provide accurate and timely information about crop locations and productivity on a global scale. This paper aims to provide an overview of how recent advances in machine learning approaches and the availability of data from earth observing satellites can dramatically improve our ability to automatically map croplands over long period and over large regions.

The remainder of paper is organized as follows. Section 2, 3, and, 4 provide the discussion for three applications where ML approaches are beginning to show great promise. For each application, we highlight the involved machine learning challenges and then discuss the proposed approaches and results. Section 5 concludes with discussion of major challenges that need to be addressed before ML approaches will reach their full potential for this problem of great societal relevance.

# 2. Cropland Mapping

Effective cropland mapping is important since it provides accurate and timely agricultural information and also helps track the consumption and transfer of water, nutrients and energy. In this task, we aim to use machine learning techniques to detect what types of crops are planted at each location in a large target region.

From early spring, satellites can capture mostly crop residues left on the ground from previous years. Then farmers start preparing to plow and fertilize the land and make it ready for seeding. They tend to grow crops in summer, and the croplands gradually turn into green as crops grow up. Finally, the greenness level will decrease after crops are harvested. Some crops have distinct growing patterns, e.g., they turn in green much faster than other crops. The availability of multi-temporal data allows modeling the temporal growing process of crops and thus can be helpful for classification.

There are several challenges in applying machine learning models to map croplands. First, different crops are very likely to be confused with each other using the data captured on a single date. An individual remote sensing image can also contain much noise due to aerosols and acquisition errors (Karpatne et al., 2016). Although many researchers have successfully applied machine learning models to map croplands using the remote sensing imagery at a single snapshot (Jia et al., 2016; Chen et al., 2011), they rely on the assumption that the imagery is available for the period during which target land covers can be identified. Also, they mostly focus on distinguishing between specific crops and cannot generalize to more complex scenarios where different crops are differentiable at different time. In contrast, it is known that crops can be better distinguished based on their temporal growing patterns. For example, previous research (Sakamoto et al., 2010) shows that corn and soybean look very similar in most dates of each year, but are differentiable at certain growing stage using their temporal patterns.

Moreover, while the multi-temporal remote sensing data cover the entire year under a regular time interval, crops show their distinctive temporal patterns only during certain period. When mapping different types of crops, we should focus on the period after seeding and before harvesting; otherwise the residues on the ground make it difficult to differentiate between crops. These periods when crops can be better identified are also referred to as the discriminative period. Therefore, it is critical to develop a classification model that can automatically pay attention to the discriminative periods while reducing the negative impact from other periods. The variability during other periods are not relevant to the detection of crops.

As mentioned earlier, temporal features that capture the growth patterns are helpful in distinguishing between different crops. While traditional machine learning models utilize the feature-based or distance-based methods to classify multi-temporal data (Xing et al., 2010), recent advances in deep learning have provided much more effective ways of modeling complex temporal dependencies, which greatly improves the classification performance.

In our research, we have developed deep learning approaches to automatically extract such temporal features from multi-temporal remote sensing data. Among all the deep learning models, the Recurrent Neural Networks (RNN) model is mostly widely used to handle temporal data. However, RNN is known to suffer from the vanishing-gradient problem (Bengio et al., 1994). Basically, it gradually loses the connection to previous time steps as time progresses.

To handle this limitation of RNNs, we use the Long-Short Term Memory (LSTM) model, which is an extension of the standard RNN model and has found tremendous success in handling sequential data in commercial applications (Luong et al., 2015; Salehinejad et al., 2017). This model can automatically extract complex long-term data dependencies from multi-temporal data and encode the extracted crop patterns into high-level feature representations.

Specifically, the LSTM model generates a high-level abstract temporal feature representation $h_t$ at each time step $t$. This representation encodes not only the information at current time step, but also the inherited temporal information from previous time steps. The hidden representation is computed in a recurrent process as $h_t = LSTM(x_t, h_{t-1})$, where the function $LSTM(\cdot)$ is defined based on the structure of the LSTM model. Prior works in the context of commercial and healthcare applications have shown that using the extracted temporal features $h_t$ for classification can achieve a much better performance than directly using the original input data (Goodfellow et al., 2016).

Next, we detect the discriminative period and use the extracted temporal features by LSTM to conduct classification. For this task, we use the attention model (Luong et al., 2015) to identify the period after seeding and before harvesting since the crop characteristics can only be captured during this period. The attention model is widely used in machine translation and image captioning (Luong et al., 2015; You et al., 2016) for its ability to automatically find a specific portion of input data that is relevant to the target output. In particular, the attention model produces an attention weight $\alpha_t$ for each time step $t$ to measure its contribution to crop classification. The higher value of $\alpha_t$ indicates that the time step $t$ contains more relevant information for identifying crops.

Next, temporal features from all the time steps are aggregated based on the obtained attention weights, as $\sum_t \alpha_t h_t$. The aggregated features contain more information from the discriminative time steps given their higher attention weights. The aggregated features are then used to conduct classification by a fully connected layer (Jia et al., 2019).

This method is implemented to classify corn vs. soybean in southwestern Minnesota, US in 2016. Our evaluations focus on these two major crops since they take over 90% area in southwestern Minnesota. More importantly, their labels provided by USDA Crop Data Layer product (USDA CDL) are more accurate than other minor crop types. We use the MODIS MODI09A1product (NASA Earthdata) as input features. This dataset provides global data for every 8 days at 500m spatial resolution. At each date, MODIS dataset provides reflectance values on 7 spectral bands for every location. To better learn short-term temporal patterns, we concatenate spectral features in every 32-days window as a time step and slide the window by 8 days. Totally we have 43 time steps in a year.

We randomly select 1,000 data points (500 corn and 500 soybean) from different locations (i.e., different MODIS pixels) in southwestern Minnesota. We predict the crop types for another set of 2,000 randomly selected locations (1,000 corn and 1,000 soybean) from southwestern Minnesota in 2016. We name our proposed method LSTM$^{ATT}$ and compare it against multiple widely used machine learning baselines, including Artificial Neural Networks (ANN), Random Forest (RF), SVM$^{HMM}$ (Altun et al., 2003), 1-NN$^{DTW}$ (Nayak et al., 2018), S2V (Xun et al., 2016), and standard LSTM. Among these baseline methods, ANN and RF are directly applied to the concatenation of multi-temporal data and thus ignore the dependencies between different dates. The remaining baselines are all commonly used sequential models. The performance is reported in Table 1 (2016-test). The method DA in Table 1 will be introduced in Section 2.2.

According to the 2016-test in Table 1, we can observe that our proposed method LSTM$^{ATT}$ outperforms other methods by a considerable margin. The comparison between LSTM and static baselines (ANN and RF) shows that the modeling of temporal profile can help detect land covers with increased accuracy. LSTM also outperforms other sequential baselines (e.g., SVM$^{HMM}$, 1-NN$^{DTW}$, S2V) because LSTM can extract representative temporal patterns by exploring complex dependencies across different spectral bands and across time. Also, the improvement from LSTM to LSTM$^{ATT}$ shows that the attention model assists in further improving the classification performance by explicitly modeling the discriminative period.

Our method LSTM$^{ATT}$ detects the discriminative period (i.e., the period with highest attention) weights is from Jun 9 to July 11. To verify this result, we show high-resolution Sentinel-2 images at 10m resolution. Fig. 1 (a) and Fig. 1 (b) show some corn and soybean patches in four example regions using Sentinel-2 images on Jun 23, which show that corn patches turn into green faster than nearby soybean patches. Such difference can also be verified by the NDVI time series as shown in Fig. 2, where corns show a much higher NDVI than soybeans at the early stage (red dashed box). It is easy to see from Sentinel images and NDVI series that these two crop types can be easily distinguished in this period.

Our method also detects the discriminative period (with higher attention weights than average) from Jul 19 to Aug 20. During this period, both corn and soybean samples show very high greenness level and therefore it is difficult to distinguish between them by human (e.g., the Aug 06 Sentinel-2 image shown in Fig. 1 (c). Here to verify that this period is indeed a discriminative period, we only use the multi-spectral features from Jul 19 to Aug 20 to train and test a simple ANN model, which produces AUC and F1-score of 0.894 and 0.806, respectively. It is noteworthy this is better than the ANN baseline that is trained using full-year sequences (AUC 0.863, F1 0.797). This improvement demonstrates that our framework has potential to detect the discriminative period from the full multi-spectrum, which cannot even be observed directly by human experts.

The successful cropland mapping also helps analyzing the distribution of winter cover crops that are planted in the autumn after harvest of the summer grain crop. Monitoring the distribution of cover crops is important since they allow for a more efficient use of resources while maintaining or improving productivity and enhancing the quality of the environment (Dabney et al., 2001; Strock et al., 2004). In addition, increasing the diversity of vegetation on the landscape has numerous benefits including weed, pest, and disease resistance; improved soil water holding capacity and nutrient cycling; pollination; and enhanced wildlife habitat (Kremen and Miles, 2012; Derksen et al., 2002; Lin, 2011).

The rate of adoption of cover crops is growing but remains small, in part because they provide limited economic benefits to farmers. The proposed method will investigate monitoring the planting of cover crops and analyze their distributions for different crop types. This will help analyze the farmers' preference in planting cover crops. Such analysis provides useful insights for government and companies to enact new polices to encourage planting cover crops.

The most widely used USDA crop data layer product does not explicitly label cover crops but either label them by the primary crops that are planted before cover crops or mistakenly mark them as ever-green vegetation such as alfalfa. A successful detection of cover crops requires analyzing the temporal profile of vegetation level and the transition patterns between primary crop types (e.g. corn and soybean) and cover crops. The most important feature for cover crops is that they stay green after the primary crops are harvested. Besides, we need to exclude the every-green vegetation such as alfalfa which do not show the same temporal patterns as primary crops in the growing season.

We show preliminary results of cover crop detection using Sentinel-2A imagery in 2016 for an area of around 2,471,054 acres around southwestern Minnesota, US. In Table 2, we report the statistics for 9 crop types in the region. We can observe that the adoption of cover crops is still limited for large crop types (0.45% for corn and 0.48% for soybean). Cover crops are planted mostly for minor crops such as spring wheat, rye, and peas. In our implementation, we find that the labeling of alfalfa is critical for accurate accounting of cover crops proportions since cover crops are rarely planted for most crop types. In Fig. 3, we show the generated cover crop map for the region.

## 3. Mapping Crops for Different Weather Conditions

Due to the high variability in the properties of different crop cover types across regions and over time, it can be challenging to build machine learning models that perform well across multiple geographies, and time periods. For example, crops can be planted under different soil types, precipitation and other weather conditions for different places and different years. Standard machine learning models only find the statistical relationships that fit the available training data. Therefore, a classification model learned from a specific year or a specific year cannot be generalized to other regions and time periods.

If the learning model can be made generalizable to different weather conditions, it can be much easier to monitor croplands globally by avoiding the need to retrain models for places or years when ground-truth labels are not available or noisy. Furthermore, such techniques can potentially help understand the changes in crop growing process across years.

For $LSTM^{ATT}$, the method proposed in Section 2 for mapping crops, performance will be degraded on two aspects. First, the classification performance will decrease since the learned statistical relationships from training data do not fit the data from different conditions. Second, the detected discriminative period becomes less accurate. The parameters in the attention model are still estimated using training data and thus the generated attention weights in testing data can be less accurate in indicating the importance of each time step.

To mitigate this issue, an ideal strategy would be to train different models for each scenario independently. However, this is challenging in many real-world applications because of a paucity of labelled data, as it is generally available only for certain regions and years.

In this work, we handle data heterogeneity through transfer learning (Pan et al., 2010) and especially domain adaptation (DA) (Jiang et al., 2008) that is extensively used in computer vision and related applications for handling data heterogeneity. Domain adaptation is a special case of transfer learning where different domains have the same feature space and class categories, but the joint probability distributions are not same

$P_S(X,Y) \neq P_T(X,Y)$ between the source domain $S$ and the target domain $T$. This situation is most common in earth observation data where same datasets are available at a global scale, but the probability distributions of the data vary due to spatial and temporal heterogeneity. Here the source domain denotes the specific regions or years with sufficient labeled data, while the target domain denotes other target regions or years of our monitoring objective.

Deep learning-based domain adaptation methods greatly outperform previous methods for domain adaptation (Tzeng et al., 2017; Ganin et al., 2016). The main reason behind their success is their ability to extract task specific features which are also consistent across different domains (Li et al., 2018; Williams et al., 2017; Venkateswara et al., 2017). Due to the data heterogeneity and paucity of labels, researchers have also applied such techniques in remote sensing applications (Tuia et al., 2016).

However, these approaches mostly focus on individual image snapshots. In contrast, several approaches have been proposed for health-care data that explore the information transfer between multi-temporal data using RNN and its variants (Purushotham et al., 2016; Yang et al., 2017; Aswolinskiy et al., 2017). All these methods leverage all the time steps equally in recurrent models to connect different domains, and thus lack the ability to avoid the transfer of non-informative time steps. Consequently, these approaches can be adversely affected by the variability in the non-informative period.

Here we discuss an approach that handles this issue by combining the discriminative period detection technique (described in Section 2.1) with the domain adaptation. As mentioned earlier, in many situations, the complete temporal duration is not required to discriminate between different categories of interest. Since farmer can switch crop types across years, adaptation on the discriminative period is especially helpful for cropland mapping because the model will not have to adapt the high variability before crops are planted.

The goal of domain adaptation (DA) is to learn a mapping from the target domain to the source domain, $g: T \rightarrow S$. This function aims to transform the data in a target domain to a distribution similar to the source domain such that the learned model can be applied to the transformed samples $g(x_T)$, where $x_T$ are the samples from the target domain. To learn the mapping between the source domain and target domain, a standard approach is to minimize the divergence between the hidden representation of two domains. Learning strategies such as adversarial deep learning have been used to minimize the divergence (Purushotham et al., 2016), which enforces that the hidden representation of the source domain cannot be distinguished from the target domain after applying the transformation $g(\cdot)$. Note that these hidden representations can potentially be extracted using the modeling framework introduced in Section 2.1.

Instead of directly conducting adversarial regularization on multi-spectral features or the hidden representation, which have been widely adopted in previous works (Purushotham et al., 2016), we apply the adversarial training on the weighted summation of hidden representation at different time steps to take account of the seasonality. In this way, the adaptation process can pay more attention to the discriminative period. Specifically, we utilize the weights obtained from the attention model described in Section 2.1, as these weights indicate the importance of each time step in classification.

However, previous research has shown that the attention model can be severely impacted when applied across different domains (Kang et al., 2018). Hence, to utilize the attention model under different weather conditions, it is important to ensure that the robustness of the attention model when applied across domains. To address this problem, we introduce another regularizer on the difference of attention weights between original samples $x_T$ and transformed samples $f(x_T)$. This mechanism has shown to be able to improve the robustness of the attention model and successfully fix the attention weights.

We implement the proposed domain adaptation (DA) method on the same training dataset with Section 2.1, which is collected in 2016. Here we test the model to two sets of data points collected in 2015 and 2011. Each test dataset contains 2,000 locations which are randomly selected from southwestern Minnesota.

According to Table 1, the performance of each method is in general degraded in 2015-test and 2011-test compared with the tests in the same condition (2016-test). The domain adaptation-based approach (DA) shows superior performance compared with LSTM$^{ATT}$ in 2015 and 2011 since it can reduce the divergence between source and target domains.

Now we assess the impact on the attention model by the data heterogeneity. Fig. 4 (a) shows the obtained attention weights for the corn locations by LSTM$^{ATT}$ in 2016 (Train) and 2015 (Test), as well as the obtained relevance scores by DA in 2015. Fig. 4 (b) shows the obtained attention weights for the corn locations by LSTM$^{ATT}$ in 2016 (Train) and 2011 (Test), as well as the obtained relevance scores by DA in 2011.

For both tests, we can observe that the LSTM$^{ATT}$ networks cannot detect a period with obviously higher attention weights when it is directly applied to the testing scenario. Therefore, it cannot precisely capture the discriminative period. In contrast, DA is capable of mitigating the impact of variability across domains and thus producing meaningful relevance scores.

The proposed method also enables interpreting the shift of discriminative period across years. For example, we can easily observe from Fig. 4 (a) that the crops in 2015 are planted earlier than the crops in 2016. This monitoring capacity is important in tracking farmers' behaviors in correspondence to weather changes. To verify this finding, we show high-resolution Landsat images at the beginning of July in 2015 (Fig. 5 (a)) and 2016 (Fig. 5 (b)). It can be seen that the selected region shows higher greenness level at this selected time in 2015 than in 2016. Hence farmers are more likely to grow crops earlier in 2015.

## 4. Early Crop Detection

Monitoring crop types and planting area is timely science that can inform prior actions for natural resources. Hence, it is of great interest to governments and companies to obtain the agricultural information in current year as soon as possible.

Traditional RNN-based classification methods generate class labels only for the entire sequence (Nayak et al., 2018; Jia et al., 2019) or for each time step separately (Jia et al., 2017; Zhang et al., 2017). However, these methods cannot be used to track the confidence over time.

Preliminary results in detecting corn, soybean and sugar beets are shown in Fig. 6. Our method captures that corn samples quickly gain confidence at the $8^{th}$ and $9^{th}$ time steps, which correspond to Jun 14~Jun 24. To validate the correctness of this finding, we show the RGB image of an example region captured on Jun 24 in Fig. 6 (a). We can clearly see that in the early growing season, corn turns into green more quickly than soybean and sugar beets, and therefore can be identified in this period.

We can also observe that sugar beets samples still gain confidence after October. We show another RGB image on Oct 05 (the same example region) in Fig. 6 (b). While corns and soybeans have been harvested, the cropland of sugar beets still remain green. These results show some promising insights that machine learning approaches can be used to track the classification confidence over time. The progression of classification confidence can potentially lead to an early detection of specific crop types when they reach sufficient confidence level.

## 5. Conclusion

As demonstrated by results in this paper, recent advances in machine learning and the availability of earth-observing satellite data can greatly improve our capability to monitor croplands over space and over time. While this technology is poised to play a key role in addressing issues related to food security at global-scale, advances are needed in many areas to realize this goal.

For example, while machine learning techniques are beginning to show success in extracting temporal patterns to create maps of crops at pixel level, these approaches (Jia et al., 2017; Jia et al., 2019; Lyu et al., 2016) have only been designed for remote sensing data that is available with high frequency, e.g. MODIS (250m, daily). However, the spatial resolution of such data is quite poor, which makes them unsuitable for monitoring small-scale farms that are quite common in many parts of the world. Higher spatial resolution data (at 10m resolution) is available from Sentinel but it is available every 5 or 10 days (depending on specific years and regions). Hence new advanced machine learning models need to be developed to combine coarse-resolution data and high-resolution data (e.g., Sentinel (10m, every 10 days in 2016) and Landsat (30m, every 16 days)) such that it can leverage the temporal patterns from low-resolution data while also mapping croplands at high resolution.

Finally, traditional machine learning and data mining algorithms fail to take advantage of the wealth of information about physical principles or human behaviors and practices that govern or have huge impact on the crop growth. For example, a unique aspect of crop yield estimation that differentiates it from standard classification or regression tasks in machine learning is that the crop growing process under specific environment is governed by relevant physical principles. Researchers in agricultural community have also built physics-based models to simulate these physical principles (Srinivasan et al., 2010; Jones et al., 2003). The ability to integrate such mechanistic knowledge in a machine learning framework will be key to advancing the state of the art in estimating crop yields.

Table 1. The performance of different machine learning models in mapping different crop types in 2016, 2015 and 2011. The training data is taken only from 2016. The performance is measured using Area Under Curve (AUC) score and F-1 measure.

|  | *2016-test* | | *2015-test* | | *2011-test* | |
| --- | --- | --- | --- | --- | --- | --- |
|  | AUC | F1 | AUC | F1 | AUC | F1 |
| ANN | 0.863 | 0.797 | 0.665 | 0.679 | 0.664 | 0.602 |
| RF | 0.863 | 0.788 | 0.672 | 0.688 | 0.662 | 0.523 |
| SVM$^{HMM}$ | 0.813 | 0.790 | 0.715 | 0.698 | 0.706 | 0.688 |
| 1-NN$^{DTW}$ | 0.814 | 0.792 | 0.703 | 0.687 | 0.700 | 0.685 |
| S2V | 0.837 | 0.806 | 0.736 | 0.712 | 0.730 | 0.706 |
| LSTM | 0.865 | 0.807 | 0.767 | 0.718 | 0.762 | 0.704 |
| LSTM$^{ATT}$ | **0.909** | **0.811** | 0.799 | 0.753 | 0.779 | 0.721 |
| DA | **0.909** | **0.811** | **0.840** | **0.775** | **0.831** | **0.758** |

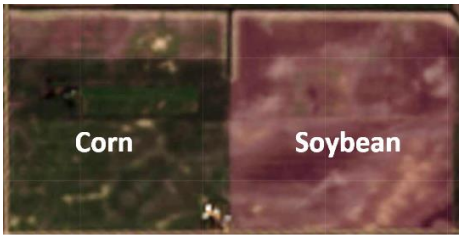

(a)

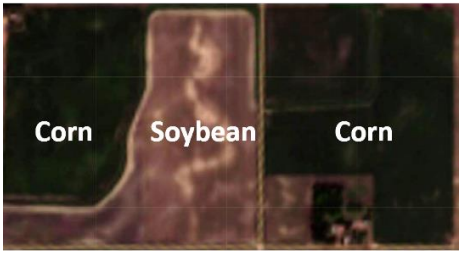

(b)

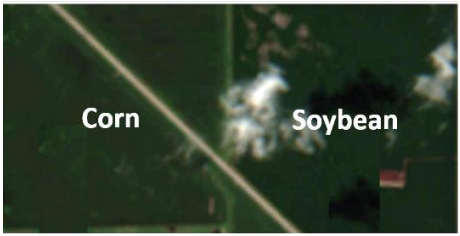

(c)

Figure 1. Sentinel-2 satellite images in RGB at 10m spatial resolution. (a) (b) Cropland patches with corn and soybean on Jun 24, 2016. Corn shows higher greenness level than soybean on this date. (c) Another cropland patch captured on Aug 06, 2016, where corn and soybean cannot be easily distinguished. Each image is approximately a 1500m×750m area.

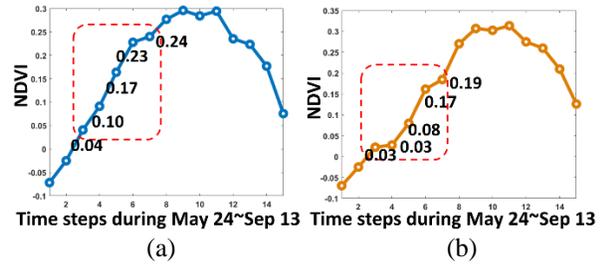

Figure 2. The average NDVI (greenness) series for (a) corn and (b) soybean during May 24 - Sep 13, 2016. The red boxes indicate the detected discriminative period Jun 9 - Jul 11, 2016.

Table 2. The distribution of different crops and the use of cover crops in our study region (southwestern Minnesota).

|  | Total (Acre) | Cover crop (%) | Cover crop (Acre) |
| --- | --- | --- | --- |
| Corn | 1014653 | 0.45 | 4597 |
| Soybean | 716215 | 0.48 | 3421 |
| Sweet corn | 35617 | 7.18 | 2556 |
| Sprint wheat | 9229 | 45.19 | 4171 |
| Rye | 561 | 62.17 | 349 |
| Oats | 523 | 40.85 | 214 |
| Sugarbeets | 71388 | 3.53 | 2519 |
| Dry beans | 14774 | 10.23 | 1511 |
| Peas | 6255 | 56.74 | 3549 |
| Total | 1869215 | 1.22 | 22888 |

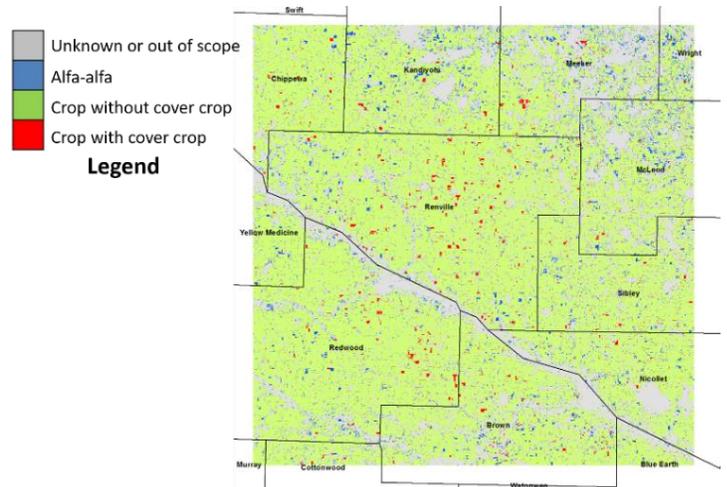

Figure 3. The generated maps for alfalfa and cover crops in our study region. This area is around 2471054 acre, or $10^8$ Sentinel pixels (at 10 m resolution).

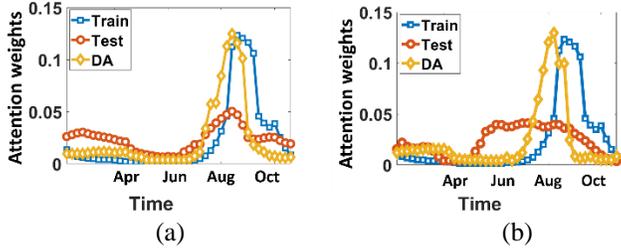

Figure 4. The impact of heterogeneity on the attention model in (a) 2015-test, and (b) 2011-test for cropland. Train: the attention weights on training data in 2016. Test: the attention weights on test data by directly applying the LSTM$^{ATT}$ model. DA: the obtained attention weights on test data using the proposed DA method.

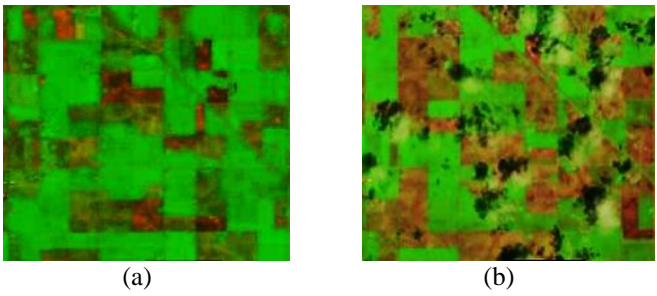

Figure 5. The Landsat images (in RGB, 30 m resolution) for an example region in southwestern Minnesota in (a) 2016 and (b) 2015 at the beginning of July.

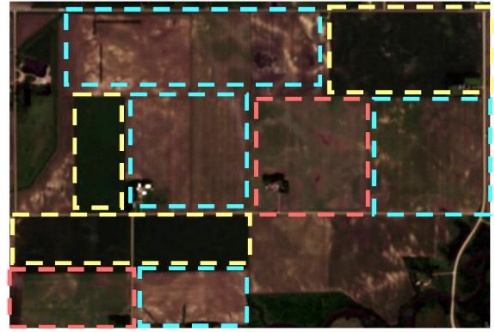

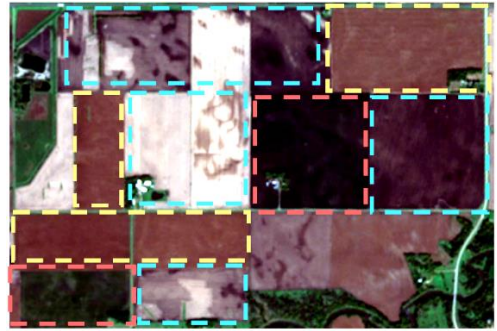

Figure 7. The Sentinel imagery in RBG captured on (a) Jun 24 and (b) Oct 05. Color legend for blocks: yellow - corn, blue - soybean, red - sugarbeet. Each area is approximately a 3000m×2000m area.

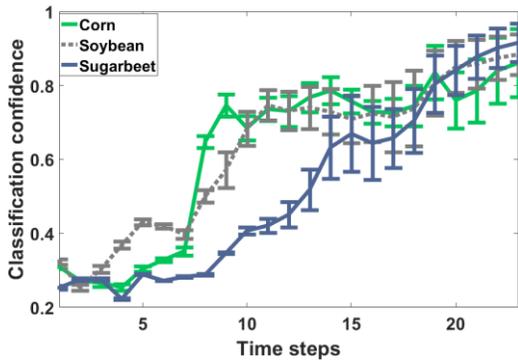

Figure 6. Confidence progression for corn, soybean and sugarbeets from Apr 05 to Nov 11 in 2016 (totally 23 time steps). The error bar represents the standard deviation.